\pdfoutput=1

\documentclass[11pt]{article}

\usepackage{acl}
\usepackage{amsmath}
\usepackage{times}
\usepackage{latexsym}
\usepackage{booktabs}
\usepackage[T1]{fontenc}

\usepackage[utf8]{inputenc}

\usepackage{microtype}
\usepackage{booktabs}
\usepackage{graphicx}
\title{Prompt Learning for Domain Adaptation in Task-Oriented Dialogue}

\author{Makesh Narsimhan Sreedhar \\
  University of Wisconsin - Madison \thanks{\hspace{0.5em} Work done as part of an internship with NVIDIA} \\
  \texttt{msreedhar@wisc.edu} \\\And
  Christopher Parisien \\
  NVIDIA \\
  \texttt{cparisien@nvidia.com}
}

\begin{document}
\maketitle
\begin{abstract}
Conversation designers continue to face significant obstacles when creating production-quality task-oriented dialogue systems. The complexity and cost involved in schema development and data collection is often a major barrier for such designers, limiting their ability to create natural, user-friendly experiences.
We frame the classification of user intent as the generation of a \textit{canonical form}, a lightweight semantic representation using natural language. We show that canonical forms offer a promising alternative to traditional methods for intent classification. By tuning soft prompts for a frozen large language model, we show that canonical forms generalize very well to new, unseen domains in a zero- or few-shot setting. The method is also sample-efficient, reducing the complexity and effort of developing new task-oriented dialogue domains.

\end{abstract}

\section{Introduction}

Task-oriented dialogue systems in Conversational AI are challenging for developers to create. The current generation of dialogue frameworks requires developers to define actions (\textit{intents}) and parameters (\textit{slots}) that the natural language understanding (NLU) module accepts. This is then used to populate API service calls that operate in the backend to fulfill the user request. Casting natural language utterances from the user to a discrete set of intents and slots is often not very intuitive. This in turn leads to a situation where developers rely on hand-crafted rule-based grammars or a large annotated set of training samples for machine learning models to implement a given design. Any change to the design of the dialogue system would then require the developers to revisit and modify the implementation which is very often a time-consuming process. In this work, we aim to make dialogue system design easier and more intuitive.  

The tremendous success of pre-trained language models such as BERT \citep{devlin-etal-2019-bert} have made them the de facto standard for most intent classification and slot-filling tasks. However, these models are not immune to the challenge of adapting and extending existing models to new domains. One adaptation approach that has exploded in popularity in recent times is the usage of prompts with these language models. With a task description and few samples showing the input-output pairs, these language models become extremely effective at solving these tasks, especially at larger model sizes.

Manually specifying prompts suffers from sensitivity to phrasing; we get widely varying results based on how we frame the prompt. Prompt tuning \citep{prompt-tuning} and p-tuning \citep{ptuning} have emerged as strong alternatives to manual prompt designing and they help optimize task-specific prompt tokens to get the best performance while keeping the language model itself frozen.    
In this work, we explore the task of intent classification using these large language models and p-tuning. Generative methods for classification tasks have not been widely adopted because generation is inherently difficult to control and utilize for further downstream tasks. Using our experiments on the Schema Guided Dialogue \citep{sgd-dataset} dataset and the Virtual Assistant Benchmark \citep{assistant-dataset}, we show that with p-tuning we can achieve promising zero-shot and few-shot generalization capabilities to unseen domains. 

In the task of intent classification, the intent labels provided as part of the dataset are usually terse and rigid. Generative models generalize better when intent labels are more descriptive but structured at the same. We borrow some aspects and terminology from semantic parsing to cast the intent labels to a more compositional format, known as \textit{canonical forms}. In the traditional sense, canonical forms are paraphrases of the user utterances to convert them to a form that the semantic parser can operate on to output logical representations. In our use case, we loosely use the term, \textit{canonical forms}, to refer to intent labels that are more descriptive than the discrete ones but are not too verbose, \textit{e.g.}, "transfer\_money" $\rightarrow$ "transfer money to bank account". We manually frame these canonical forms and do not rely on any grammar, simplifying the approach.

We observe that using such \textit{canonical forms} as labels for the intent classification task allows the model to generalize better to domains that are adjacent, but not seen at train time (\textit{e.g.}, \textit{Flight Reservations} $\rightarrow$ \textit{Bus Bookings}). We also find that it is beneficial to do a two-stage P-tuning for domain adaption, \textit{i.e.}, once we have a p-tuned large language model on a wide set of domains, we can continue p-tuning this model on a small set of labelled samples from the target domain to allow the model to generalize better. We find that this few-shot approach works very well and this has promising implications for developers for dialogue systems; with minimal effort it would be feasible to adapt an existing model pre-trained on multiple domains to a new domain. In summary, our contributions are:

\begin{itemize}
\item We cast the problem of intent classification into a generative approach and rewrite intent labels in a more descriptive format (\textit{canonical forms}).
    \item When using such canonical forms, generative approaches with Large Language Models(LLMs) show promising results when compared with traditional methods for intent classification.
\item Generative models generalize very well to unseen domains in zero-shot and few-shot settings when compared with BERT-style approaches. 
\item We demonstrate the sample efficiency of p-tuning LLMs where we can achieve close to full dataset performance with a fraction of the data.

\end{itemize}

\section{Method}

In this section, we describe the creation of canonical forms and the prompt tuning technique we adopt for intent classification in the task-oriented dialogue setting.

\subsection{Canonical Forms}
\label{sec:canonical}
Canonical forms are usually paraphrases of the user utterance to a standardized form that can be utilized by downstream systems. These forms are traditionally obtained by using a set of grammar rules written by experts. The output of this process is a natural language sequence, but structured in a form that makes it better suited for a semantic parser. Different semantic parsers employ different canonical forms and thus transfer across datasets is quite challenging.

\begin{table}[h]
\resizebox{\columnwidth}{!}{%
\begin{tabular}{@{}ll@{}}
\toprule
Utterance                             & Canonical Form                                      \\ \midrule
what is the newest published article? & article that has the largest publication date       \\
who has published the most articles?  & person that is author of the most number of article \\ \bottomrule
\end{tabular}}
\caption{Examples of canonical forms corresponding to user utterances from the Overnight \citep{wang-etal-2015-building} semantic parsing dataset.}
\end{table}

Our work uses canonical forms as a method of obtaining the intent of a user utterance. Traditionally, intent labels tend to be terse, which makes it difficult for models to generalize to unseen domains. The expressive and compositional nature of language models can be exploited if the intent labels are more verbose, allowing them to extrapolate the generated intents to capture even novel domains. At the same time, if the intent labels tend to be very long and riddled with descriptions, the language models become susceptible to hallucinations. Our work proposes the use of canonical forms as a way of establishing a balance between being terse and too verbose. We map intent labels to short descriptive phrases, \textit{e.g.}, "\textit{check\_balance}" $\rightarrow$ "\textit{check balance in bank account}". Unlike traditional canonical forms, we do not use any formal grammar to perform this mapping and the phrases are manually specified. We believe that such an approach would reduce the burden on developers and designers of conversational systems.  

\subsection{P-tuning}
Large Language Models (LLMs) have exhibited remarkable generalization capability when queried using \textit{prompts} that contain examples of the task to be performed. However, the performance of LLMs varies widely depending on how such prompts are constructed. In order to overcome this issue of LLM sensitivity to the format of the prompt, multiple studies have come up with methods for automated prompt construction using discrete tokens \citep{prompt-tuning} as well as soft tokens \citep{ptuning}. 

In this work, we utilize the p-tuning approach that appends learnt \textit{soft} tokens into the prompt that is fed to the LLM. The soft tokens traditionally do not have a mapping to words/subwords in the model vocabulary and are simply vectors optimized using gradient descent. Following the setup proposed by \citet{ptuning}, we use an LSTM model to learn and predict these soft tokens. The parameters of the LLM are frozen and only the parameters of this LSTM model are updated during p-tuning. We initialize the LSTM with random weights at the beginning of the p-tuning process and then update it during the training stage to output the optimal soft tokens. At the end of the training phase, we store these soft tokens and append them with the prompt to the LLM to get its prediction. The advantage of p-tuning is that we freeze the LM weights and update only the weights of the LSTM (14M parameters). This results in modifying only a very small fraction of the weights compared to traditional finetuning where all of the weights are updated.     

The LM of choice in our experiments are the Megatron-GPT \citep{megatrongpt} models that are \textit{decoder-only} transformers.

\section{Experimental Setting}
In this section, we describe the datasets used, the baselines we use for comparison and the evaluation metrics.

\subsection{Datasets}
\label{sec:datasets}
We consider two widely known datasets in the dialogue community, the Schema Guided Dialogue (SGD) dataset \citep{sgd-dataset} and the Virtual Assistant dataset \citep{assistant-dataset}.\\\\
\textbf{Schema Guided Dialogue} - This dataset covers 16 domains and has over 16k annotated conversations. The domains span a variety of user actions, including setting calendars and alarms, travel booking (car rentals, flights, buses and trains), music, weather, movies, and more. The dataset also contains \textit{multi-domain} dialogues where the utterances switch between domains. For the purpose of our experiments, we consider only the \textit{single-domain} dialogues with 37 intents across all utterances. \\\\
\textbf{Virtual Assistant Dataset} - This dataset covers 21 domains with 64 intents across all utterances. As the name suggests, the domains relate to user queries over a wide range of topics, including operating smart-home devices, media consumption, weather and travel. It has over 25k annotated user utterances that identify intents and slot values.

\subsection{Prompt Template}

The prompts that we use for intent classification have the following format 

\begin{align}
    <v_1..v_n> \quad {utterance} \quad intent:{canonical}
\end{align}

where $<v_1, v_2, .., v_n>$ indicate the virtual tokens.

During the training stage of p-tuning, the model is shown the entire sequence, but the loss is computed only on the \textit{answer} which in this case is the predicted canonical form. During inference, the context to the model includes the sequence until the word \textit{"intent:"} and the model completes the sequence with its prediction for the intent. We use 100 virtual tokens with our prompt-encoder being an LSTM model with 2 layers.

\subsection{Evaluation Method}

\textbf{Intent Classification} Evaluating generative models for a classification task is not straightforward. This is further complicated by the fact that our model generates a canonical form identifying the intent of a given user utterance. We propose two methods to cast this generation problem to a classification setting. The difficulty arises from the fact that generated sequences very often differ from the exact gold truth sequence that the model sees as part of training. We utilize two approaches based on associating the generated canonical form to its closest label, \textit{i.e.}, a nearest neighbor search. Once the canonical form label has been identified as the prediction, it becomes trivial to compute the classification accuracy. Since we already have a one-to-one mapping between canonical form labels and the discrete intent label, we can easily measure the performance of our model.

\begin{itemize}
    \item Using Fasttext Embeddings \citep{bojanowski2016enriching}: We take the mean of all the embedding vectors of the generated canonical form and consider the vector obtained to be the representation of the whole sequence. We compute similar vectors for all the canonical form labels and consider the canonical form label that has the maximum cosine similarity with the generated one as the model's prediction. 
    
    \item Using Sentence Transformers \citep{reimers-2019-sentence-bert}: We use the \textit{miniLM-QA} \citep{minilm} transformer model that has been pretrained on multiple datasets on the text entailment/semantic search task, \textit{i.e.}, given a query and a set of keys (documents/labels), it ranks the keys in order of relevance. We give as input to the model the generated canonical form (query) and the list of canonical form labels (keys). The model then returns the closest canonical form label to the generated canonical form which we consider as the prediction.  
\end{itemize}

\subsection{Baselines}

We consider the following baselines for the intent classification task.

\begin{itemize}
    \item \textbf{BERT-based finetuned model} (Intent Classification): We finetune BERT models on the datasets described in section \ref{sec:datasets}. While some of the Megatron-GPT models we use are larger than the BERT model in terms of number of parameters, it should be noted that the LM parameters are frozen during the training stage of p-tuning and only the weights of the LSTM (~14M parameters) are updated.
\end{itemize}

\subsection{Evaluation Settings}
We evaluate the performance of our model in two settings: in-domain and out-of-domain.

\subsubsection{In-Domain}
This setting corresponds to the traditional dataset splits where the train and test sets come from similar distributions. We p-tune the Megatron-GPT models on the train set and evaluate them on the test set for intent classification.

\subsubsection{Out-of-Domain}

In this setting, we aim to explore the generalization capability of LLMs. We hold out certain domains from the train set and use utterances from the held out domains as our test. This helps us understand how well these LLMs can generalize to unseen domains. The held out sets that we consider are:

\begin{itemize}
    \item \textbf{Schema Guided Dialogue} (SGD): We hold out utterances corresponding to \textit{bus bookings} and \textit{hotel reservations} to form our test set. The train set includes utterances from adjacent domains: flight booking and restaurant reservations. This should be a relatively easy setting for the language model to generalize to.
    \item \textbf{Virtual Assistant}: To make things more challenging, we hold out utterances corresponding to \textit{operating IOT devices} and \textit{media consumption commands} (\textit{e.g.}, commands that are variants of "play" - play movie, play audiobook). The train set does not have utterances from similar domains and this setting is more challenging for the model.
\end{itemize}

We consider the generalization capability of the model in two modes: 
\begin{itemize}
    \item \textbf{Zero-shot}: P-tune the model on the train set and evaluate zero-shot on the unseen domain test set.
    \item \textbf{Few-shot}: After p-tuning on the train set, we do a second stage p-tuning on a set of \textit{k} samples from the target domain. Unless otherwise noted, \textit{k} here is 5, 10, 50 or 100 samples. 
\end{itemize}

The few-shot paradigm may be very useful for dialogue system developers in a limited-resource setting. Developers can implement new domains using existing language models and a small set of curated examples, without the burden and expense of retraining or providing a large number of labelled samples. 

\section{Results}
In this section, we review the quantitative performance of the models for intent classification.

\subsection{Intent Classification}
We compute and list the accuracy of the baselines and our p-tuned GPT model in identifying the intent given the user utterance. 

\subsubsection{In-domain}
We find that both the p-tuned GPT model as well as the BERT baseline perform very well on the standard in-domain split where both the train and test set come from the same distribution (Table \ref{tab:indomain}). The classification accuracy of Megatron-GPT increases as we increase the model size. The trend of results remains consistent for both the SGD and Assistant datasets. 

\begin{table}[h]
\centering
\begin{tabular}{@{}lcc@{}}
\toprule
Model              & SGD  & Assistant \\ \midrule
BERT-Large         & 0.88 & 0.91      \\\midrule
Megatron-GPT - 345M & 0.87 & 0.88      \\
Megatron-GPT - 1.3B & 0.91 & 0.92      \\ 
Megatron-GPT - 5B   & 0.95 & 0.94      \\ \bottomrule
\end{tabular}%
\caption{Classification Accuracy on test sets of the SGD and Assistant datasets}
\label{tab:indomain}
\end{table}

\subsubsection{Out-of-Domain}

The out-of-domain setting is where the advantage of using a LLM becomes apparent. It is not feasible to expect a finetuned BERT model to generalize to an unseen domain not present in the train set. Such models continue to predict that the intent belongs to one of the intent labels they see during training. The p-tuned Megatron-GPT models, on the other hand, show impressive zero-shot and few-shot generalization capabilities on the SGD dataset (Table \ref{tab:ood-sgd}). For instance, having seen intents such as \textit{"buy flight roundtrip tickets"} when presented with utterances for \textit{Flight Reservations} in training, we can expect the model to reasonably generalize to utterances from \textit{Bus Reservations} with utterances like "Get me a return trip on the bus" with the model's prediction for the intent being \textit{"buy bus roundtrip tickets"}.
\begin{table}[!htbp]
\resizebox{\columnwidth}{!}{%
\begin{tabular}{@{}lllllll@{}}
\toprule
Mode            & \multicolumn{3}{l}{Bus Booking} & \multicolumn{3}{l}{Hotel Reservation} \\ \midrule
                & 345M      & 1.3B     & 5B       & 345M        & 1.3B       & 5B         \\ \midrule
Zero Shot       & 0.755     & 0.762    & 0.787    & 0.379       & 0.448      & 0.467      \\
FS - 10 samples & 0.907     & 0.789    & 0.942    & 0.793       & 0.720       & 0.939      \\
FS - 50 samples & 0.953     & 0.965    & 0.975    & 0.957       & 0.968      & 0.970    \\ \bottomrule     
\end{tabular}%

}
\caption{Zero-shot and Few Shot (FS) performance on the held out domains of the SGD dataset. The columns indicate the size of the Megatron-GPT model.}
\label{tab:ood-sgd}
\end{table}

In the Assistant dataset, the p-tuned models face the same issue as the BERT models: they struggle to generalize to completely unseen domains and the performance is close to random (Table \ref{tab:ood-asst}). Unlike in SGD, the held-out domains do not have sufficiently similar domains in training from which to generalize. However, the few-shot setting holds promise as the performance of the models improves with few samples. Since the held out domains have far more intents compared to the held out domains from the SGD dataset, we employ stratified sampling to ensure that the few-shot examples are representative of all intents in the domain. 

\begin{table}[!htbp]
\resizebox{\columnwidth}{!}{%
\begin{tabular}{@{}lllllll@{}}
\toprule
Mode                    & \multicolumn{3}{l}{IOT devices} & \multicolumn{3}{l}{Media Consumption} \\ \midrule
                        & 345M      & 1.3B     & 5B       & 345M        & 1.3B       & 5B         \\ \midrule
Zero Shot               & 0.096     & 0.011    & 0.022    & 0.037       & 0.008      & 0.012      \\
FS - 10 samples & 0.62      & 0.71     & 0.75     & 0.58            & 0.62           & 0.68            \\
FS - 50 samples & 0.69     & 0.83     & 0.87     & 0.67            & 0.86           & 0.89          \\ \bottomrule     
\end{tabular}%
}
\caption{Zero-shot and Few Shot (FS) performance on the held out domains of the Assistant dataset. The columns indicate the size of the Megatron-GPT model.}
\label{tab:ood-asst}
\end{table}
\section{Discussion}
The results on zero-shot and few-shot settings for unseen domains demonstrate that p-tuning a LLM to have intents that are more verbose than discrete labels can be very helpful. 

In this section, we analyze the impact of the structure of canonical forms, what helps the language model generalize, how sample efficient are these language models and what all this means for a developer of chatbots and dialogue systems.

\subsection{How important is framing the right canonical form?}
The phrasing of canonical forms has a significant impact on zero-shot cross domain generalization. In our initial experiments, we observed that the language models, especially the smaller ones, sometimes rely on spurious correlations to predict the intent. For instance, if the intent \textit{SearchFlightOneWay} is mapped to the canonical form \textit{search tickets for flight one way}, the model correlates the word \textit{ticket} in both the user utterance and canonical form to identify the intent. When we use this model to predict the intent of user utterances related to \textit{bus bookings} in a zero-shot manner, the model predicts that that the intent is related to a \textit{flight booking} as most utterances in the \textit{bus domain} contain the word \textit{ticket}. 

\begin{table}[h]
\centering
\begin{tabular}{@{}lccc@{}}
\toprule
\textbf{Mode }         & \multicolumn{3}{c}{\textbf{Accuracy}} \\ \midrule
              & \textbf{345M }    & \textbf{1.3B    }&\textbf{ 5B }     \\ \midrule
ZS - Original & 0.08     & 0.13    & 0.21    \\ 
ZS - Modified & 0.755    & 0.762   & 0.787   \\ \bottomrule
\end{tabular}
\caption{Zero shot (ZS) performance on utterances from \textit{Bus Bookings}. \textbf{Original} refers to having the canonical form for flight bookings as \textit{search tickets for flight one way} which led to incorrect generalizations. \textbf{Modified} refers to having the improved canonical form for flight bookings as \textit{search for flights one way}.  }
\label{tab:ood-sgd-original}
\end{table}

Rephrasing the canonical form for the intent \textit{SearchFlightOneWay} to \textit{search for flights one way} helps the model to avoid making the spurious correlation and the performance in the zero-shot setting (Table \ref{tab:ood-sgd-original}) is significantly improved.

\begin{table}[h]
\centering
\begin{tabular}{@{}lccc@{}}
\toprule
\textbf{Mode }         & \multicolumn{2}{c}{\textbf{Accuracy}} \\ \midrule
              & \textbf{345M }    & \textbf{1.3B    }     \\ \midrule
ZS - Original & 0.08     & 0.13       \\ 
FS 10 samples- Original & 0.76     & 0.72       \\ 
FS 20 samples - Original & 0.84    & 0.87     \\ \bottomrule
\end{tabular}
\caption{Zero shot (ZS) performance on utterances from \textit{Bus Bookings}. \textbf{Original} refers to having the canonical form for flight bookings as \textit{search tickets for flight one way} which led to incorrect generalizations. Adding a small number of examples resolves the error.}
\label{tab:ood-sgd-original-fs}
\end{table}

However, the few-shot setting (Table \ref{tab:ood-sgd-original-fs}) alleviates this problem of sensitivity of the model to the canonical form structure. When we provide the model with a few samples from the the target domain, it learns to associate that the important words to distinguish between the domains are \textit{flight} and \textit{bus} and not \textit{ticket}.

\subsection{What do good canonical forms looks like?}

Based on our experiments, a set of good canonical forms has the following properties:
\begin{itemize}
    \item \textbf{Similarity in structure}: Use similar verbs for similar actions/domains, \textit{e.g.}, \textbf{book} a flight, \textbf{book} bus tickets, \textbf{search} for hotels, \textbf{search} for restaurant reservations.
    \item \textbf{ Compositional}: Using similar structures for canonical forms in similar domains naturally lends to compositionality. This makes it easier for the model to generalize in the zero-shot/few-shot setting while still allowing the developers to easily map the generations to a supported service on the backend.
    \item \textbf{Looks like natural language}: Since LLMs are pretrained on very large corpora of natural language, the benefit of pre-training is realized when the canonical forms resemble natural language rather than complex semantic forms. Making discrete intents look more like typical verb phrases brings out the expressive nature of language models. 
\end{itemize}

Future work will explore and refine methods to automate the creation of canonical forms.

\subsection{Do we need the entire training set for p-tuning?}

We look for the fewest labelled samples for p-tuning needed to get an accuracy close to accessing the entire train set. We randomly sample \textit{k} samples per intent ($k \in {5, 10, 20, 30}$) to form the train set the model is p-tuned on, and evaluate on the same test set as above. The train and test sets are from the in-domain setting for both SGD (Table \ref{tab:sgd-fs-indomain}) and Assistant (Table \ref{tab:asst-fs-indomain}) datasets.

\begin{table}[h]
\centering
\resizebox{\columnwidth}{!}{%
\begin{tabular}{@{}l|c|ccc@{}}
\toprule
\textbf{\#Samples/Intent} & \textbf{Train Size} & \multicolumn{3}{c}{\textbf{Accuracy}} \\ \midrule
                          && 345M        & 1.3B        & 5B        \\\midrule
10                         & 370 & 0.77       & 0.81       & 0.827      \\
20                        & 740 & 0.82        & 0.83        & 0.844      \\
30                        & 1110 & 0.84        & 0.85        & 0.87     \\\bottomrule
\end{tabular}%
}
\caption{Accuracy on the SGD test set when using only k samples per intent. The columns indicate the size of the Megatron-GPT model used.}
\label{tab:sgd-fs-indomain}
\end{table}

\begin{table}[h]
\centering
\resizebox{\columnwidth}{!}{%
\begin{tabular}{@{}l|c|ccc@{}}
\toprule
\textbf{\#Samples/Intent} & \textbf{Train Size} & \multicolumn{3}{c}{\textbf{Accuracy}} \\ \midrule
                          && 345M        & 1.3B        & 5B        \\\midrule
10                        &640& 0.69        & 0.81        & 0.84      \\
20                        &1280& 0.74        & 0.84        & 0.91     \\
30                        &1920&  0.79        &0.87         & 0.91     \\ \bottomrule
\end{tabular}%
}
\caption{Accuracy on the Assistant test set when using only k samples per intent. The columns indicate the size of the Megatron-GPT model used.}
\label{tab:asst-fs-indomain}
\end{table}

\subsubsection{Comparison with BERT}

We observe that Megatron-GPT is more sample efficient than BERT-type models, even when adjusting for the number of parameters. We use the 345M parameter version of the Megatron-GPT for a fair comparison. We finetune BERT-Large and p-tune the GPT model on the same training subset of the SGD dataset. Results are shown in Table \ref{tab:gptvsbert}.

With a small number of samples (10 per intent), both Megatron-GPT and BERT-Large have very similar performance. But with small increases in the number of labelled samples per intent in the train set, we observe that the performance of the GPT model improves faster than the BERT model.

\begin{table}[h]
\centering
\begin{tabular}{@{}l|c|ccc@{}}
\toprule
\textbf{\#Samples/Intent} &  \multicolumn{2}{c}{\textbf{Accuracy}} \\ \midrule
                          & 345M        & BERT                \\\midrule
10                        & 0.77        & 0.75             \\
20                        & 0.82        & 0.767             \\
30                        &  0.84        &0.773           \\ \bottomrule
\end{tabular}%

\caption{Accuracy on the SGD test set when using only k samples per intent. MegatronGPT-345M is more sample efficient than BERT-Large.}
\label{tab:gptvsbert}
\end{table}



\subsection{What does this mean for dialogue system developers?}

Task-oriented dialogue systems are challenging to create. Most common frameworks cast utterances into discrete intents and slots, but it is often not clear how to define these concepts for a given design. Such frameworks also employ NLU models that often require the creation of either rule-based grammars or a significantly large corpus of labelled samples. While ML-based approaches have come a long way, distributional shifts in the way utterances are structured can degrade performance.
By leveraging LLMs, our approach reduces the effort involved in framing intents and training classifiers.
Because of the flexibility in canonical form schemas and the sample efficiency of p-tuning, we argue that development of new task-oriented dialogues becomes simpler and faster. We envision a setting where a model publisher trains and releases a general-purpose p-tuned language model covering a broad set of cases. A conversation designer may then write a small set of example queries, submit a brief p-tuning job, and deploy a new application with minimal cost.


\section{Conclusion}
We explore the use of Large Language Models and p-tuning for intent classification in task-oriented dialogue systems. We show framing intent labels into more verbose forms allows LMs to exploit the underlying structure better and exhibit impressive zero-shot and few-shot generalization. We also analyze how important the phrasing of the verbose forms are and how many samples are needed to get good quantitative performance. We hope that this work on using sample efficient LLMs serves to motivate further research in making ToD systems simpler and quicker to develop.

\section{Acknowledgements}

The authors would like to thank Zhilin Wang, Virginia Adams, Sandeep Subramanian, Vlad Getselevich, Prasoon Varshney, and Jonathan Cohen for many useful discussions during the course of this work.

\bibliography{anthology,custom}

\begin{thebibliography}{10}
\expandafter\ifx\csname natexlab\endcsname\relax\def\natexlab#1{#1}\fi

\bibitem[{Bojanowski et~al.(2016)Bojanowski, Grave, Joulin, and
  Mikolov}]{bojanowski2016enriching}
Piotr Bojanowski, Edouard Grave, Armand Joulin, and Tomas Mikolov. 2016.
\newblock Enriching word vectors with subword information.
\newblock \emph{arXiv preprint arXiv:1607.04606}.

\bibitem[{Devlin et~al.(2019)Devlin, Chang, Lee, and
  Toutanova}]{devlin-etal-2019-bert}
Jacob Devlin, Ming-Wei Chang, Kenton Lee, and Kristina Toutanova. 2019.
\newblock \href {https://doi.org/10.18653/v1/N19-1423} {{BERT}: Pre-training of
  deep bidirectional transformers for language understanding}.
\newblock In \emph{Proceedings of the 2019 Conference of the North {A}merican
  Chapter of the Association for Computational Linguistics: Human Language
  Technologies, Volume 1 (Long and Short Papers)}, pages 4171--4186,
  Minneapolis, Minnesota. Association for Computational Linguistics.

\bibitem[{Lester et~al.(2021)Lester, Al-Rfou, and Constant}]{prompt-tuning}
Brian Lester, Rami Al-Rfou, and Noah Constant. 2021.
\newblock \href {https://doi.org/10.48550/ARXIV.2104.08691} {The power of scale
  for parameter-efficient prompt tuning}.

\bibitem[{Liu et~al.(2021)Liu, Zheng, Du, Ding, Qian, Yang, and Tang}]{ptuning}
Xiao Liu, Yanan Zheng, Zhengxiao Du, Ming Ding, Yujie Qian, Zhilin Yang, and
  Jie Tang. 2021.
\newblock \href {https://doi.org/10.48550/ARXIV.2103.10385} {Gpt understands,
  too}.

\bibitem[{Liu et~al.(2019)Liu, Eshghi, Swietojanski, and
  Rieser}]{assistant-dataset}
Xingkun Liu, Arash Eshghi, Pawel Swietojanski, and Verena Rieser. 2019.
\newblock \href {https://doi.org/10.48550/ARXIV.1903.05566} {Benchmarking
  natural language understanding services for building conversational agents}.

\bibitem[{Narayanan et~al.(2021)Narayanan, Shoeybi, Casper, LeGresley, Patwary,
  Korthikanti, Vainbrand, Kashinkunti, Bernauer, Catanzaro, Phanishayee, and
  Zaharia}]{megatrongpt}
Deepak Narayanan, Mohammad Shoeybi, Jared Casper, Patrick LeGresley, Mostofa
  Patwary, Vijay~Anand Korthikanti, Dmitri Vainbrand, Prethvi Kashinkunti,
  Julie Bernauer, Bryan Catanzaro, Amar Phanishayee, and Matei Zaharia. 2021.
\newblock \href {https://doi.org/10.48550/ARXIV.2104.04473} {Efficient
  large-scale language model training on gpu clusters using megatron-lm}.

\bibitem[{Rastogi et~al.(2019)Rastogi, Zang, Sunkara, Gupta, and
  Khaitan}]{sgd-dataset}
Abhinav Rastogi, Xiaoxue Zang, Srinivas Sunkara, Raghav Gupta, and Pranav
  Khaitan. 2019.
\newblock \href {https://doi.org/10.48550/ARXIV.1909.05855} {Towards scalable
  multi-domain conversational agents: The schema-guided dialogue dataset}.

\bibitem[{Reimers and Gurevych(2019)}]{reimers-2019-sentence-bert}
Nils Reimers and Iryna Gurevych. 2019.
\newblock \href {http://arxiv.org/abs/1908.10084} {Sentence-bert: Sentence
  embeddings using siamese bert-networks}.
\newblock In \emph{Proceedings of the 2019 Conference on Empirical Methods in
  Natural Language Processing}. Association for Computational Linguistics.

\bibitem[{Wang et~al.(2020)Wang, Wei, Dong, Bao, Yang, and Zhou}]{minilm}
Wenhui Wang, Furu Wei, Li~Dong, Hangbo Bao, Nan Yang, and Ming Zhou. 2020.
\newblock \href {https://doi.org/10.48550/ARXIV.2002.10957} {Minilm: Deep
  self-attention distillation for task-agnostic compression of pre-trained
  transformers}.

\bibitem[{Wang et~al.(2015)Wang, Berant, and Liang}]{wang-etal-2015-building}
Yushi Wang, Jonathan Berant, and Percy Liang. 2015.
\newblock \href {https://doi.org/10.3115/v1/P15-1129} {Building a semantic
  parser overnight}.
\newblock In \emph{Proceedings of the 53rd Annual Meeting of the Association
  for Computational Linguistics and the 7th International Joint Conference on
  Natural Language Processing (Volume 1: Long Papers)}, pages 1332--1342,
  Beijing, China. Association for Computational Linguistics.

\end{thebibliography}
\bibliographystyle{acl_natbib}

\end{document}